\pdfoutput=1

\documentclass[11pt]{article}
\usepackage{authblk}
\usepackage{acl}

\usepackage{times}
\usepackage{latexsym}

\usepackage[T1]{fontenc}

\usepackage[utf8]{inputenc}

\usepackage{microtype}

\usepackage{multirow}
\usepackage{graphicx}
\usepackage{subcaption}
\usepackage{amsmath}

\begin{document}\makeatletter
\setlength\topmargin{0.0in} \setlength\oddsidemargin{-0.0in}
\setlength\textheight{9.5in} \setlength\textwidth{6.25in}
\setlength\columnsep{0.2in}
\setlength\titlebox{2.5in}
\setlength\headheight{0pt}   \setlength\headsep{0pt}
\setlength\footskip{0pt}
\thispagestyle{empty}      \pagestyle{empty}
\flushbottom \twocolumn \sloppy

\def\maketitle{\par
 \begingroup
   \def\thefootnote{\fnsymbol{footnote}}
   \def\@makefnmark{\hbox to 0pt{$^{\@thefnmark}$\hss}}
   \twocolumn[\@maketitle] \@thanks
 \endgroup
 \setcounter{footnote}{0}
 \let\maketitle\relax \let\@maketitle\relax
 \gdef\@thanks{}\gdef\@author{}\gdef\@title{}\let\thanks\relax}
\def\@maketitle{\vbox to \titlebox{\hsize\textwidth
 \linewidth\hsize \vskip 0.125in minus 0.125in \centering
 {\Large\bfseries \@title \par} \vskip 0.2in plus 1fil minus 0.1in
 {\def\and{\unskip\enspace{\rmfamily and}\enspace}%
  \def\And{\end{tabular}\hss \egroup \hskip 1in plus 2fil
           \hbox to 0pt\bgroup\hss \begin{tabular}[t]{c}\bfseries}%
  \def\AND{\end{tabular}\hss\egroup \hfil\hfil\egroup
          \vskip 0.25in plus 1fil minus 0.125in
           \hbox to \linewidth\bgroup\large \hfil\hfil
             \hbox to 0pt\bgroup\hss \begin{tabular}[t]{c}\bfseries}
  \hbox to \linewidth\bgroup\large \hfil\hfil
    \hbox to 0pt\bgroup\hss \begin{tabular}[t]{c}\bfseries\@author
                            \end{tabular}\hss\egroup
    \hfil\hfil\egroup}
  \vskip 0.2in plus 2fil minus 0.1in
}}
\makeatother
\title{Zero- and Few-Shots Knowledge Graph Triplet Extraction with\\Large Language Models}

\author[1]{Andrea Papaluca}
\author[2]{Daniel Krefl}
\author[1]{Sergio Mendez Rodriguez}
\author[3,4]{\\Artem Lensky}
\author[1, 5, 6]{Hanna Suominen}

\affil[1]{\footnotesize{School of Computing, The Australian National University, Canberra, ACT, Australia}}

\affil[2]{\footnotesize{Independent}}

\affil[3]{\footnotesize{School of Engineering and Technology, The University of New South Wales, Canberra, ACT, Australia}}

\affil[4]{\footnotesize{School of Biomedical Engineering, Faculty of Engineering, The University of Sydney, Sydney, NSW, Australia}}

\affil[5]{\footnotesize{School of Medicine \& Psychology, The Australian National University, Canberra, ACT, Australia}}

\affil[6]{\footnotesize{Department of Computing, University of Turku, Turku, Finland}}

\maketitle
\begin{abstract}
In this work, we tested the Triplet Extraction (TE) capabilities of a variety of Large Language Models (LLMs) of different sizes in the Zero- and Few-Shots settings. In detail, we proposed a pipeline that dynamically gathers contextual information from a Knowledge Base (KB), both in the form of context triplets and of (sentence, triplets) pairs as examples, and provides it to the LLM through a prompt. The additional context allowed the LLMs to be competitive with all the older fully trained baselines based on the Bidirectional Long Short-Term Memory (BiLSTM) Network architecture. 
We further conducted a detailed analysis of the quality of the gathered KB context, finding it to be strongly correlated with the final TE performance of the model. In contrast, the size of the model appeared to only logarithmically improve the TE capabilities of the LLMs. 
\end{abstract}

\section{Introduction}

The task of Triplet Extraction (TE)~\citep{te_survey} is of fundamental importance for Natural Language Processing (NLP). This is because the core meaning of a sentence is usually carried by a set of $(subject,\,predicate,\,object)$ triplets. 
Therefore, the capability to identify such triplets is a key ingredient for being able to understand the~sentence.

Currently, the State-Of-The-Art (SOTA) for TE is achieved by models that approach the TE task in an {\it end-to-end} fashion~\citep{zheng-etal-2017-joint,zeng-etal-2018-extracting,fu-etal-2019-graphrel,zeng-etal-2019-learning,tang-etal-2022-unirel}. That is, they are trained to perform all the TE sub-tasks, namely, Named Entity Recognition (NER~\citep{ner-survey}), Entity Linking (EL~\citep{el-survey}), and Relation Extraction (RE~\cite{re-survey}), together. These SOTA models follow the classic NLP paradigm, {\it i.e.}, they are trained by supervision on specific TE datasets. However, this dependence on labeled data restricts their generality and, therefore, limits the applicability of such models to the real world.

While several labeled datasets for the TE task are publicly available \cite{nyt,webnlg}, these cover only part of the spectrum of possible entities and relations. This means that a supervised model trained on this public data will be restricted to the closed set of entities and relations seen during training, implying that it may lack generalization capabilities. Producing a tailored dataset for training a model for particular applications, is, however, in general expensive \cite{johnson-etal-2018-predicting}.  

For this reason, the recent language understanding and reasoning capabilities demonstrated by Large Language Models (LLMs), such as the Generative Pre-trained Transformer 4 (GPT-4)~\citep{gpt4}, LLM Meta AI (LLaMA)~\citep{llama}, and Falcon~\citep{falcon} to name a few, have led researchers~\citep{chia-etal-2022-relationprompt,kim2023zeroshot,Wadhwa2023RevisitingRE,wei2023zeroshot,zhu2023llms} to investigate whether they represent a viable option to overcome the limitations imposed by supervised models for TE. In detail, the new approach being that at inference time the LLMs are prompted to extract the triplets contained in a sentence, while being provided with only a few labeled examples (or no example at all in the Zero-Shot setting). This LLM approach largely limits the amount of data needed to perform the task, and, in particular, lifts the restriction of adhering to a predefined closed set of relations.
However, the investigations so far indicated that the Zero and Few-Shots performance of the LLMs appears to be often underwhelming compared to the classic fully trained NLP models.

In order to enhance the abilities of LLMs in the TE task, we propose in this work to aid them with the addition of a Knowledge Base (KB). We demonstrate that augmenting LLMs with KB information, {\it i.e.}, dynamically gathering contextual information from the KB, largely improves their TE capabilities, thereby making them more competitive with classic NLP baselines. In particular, we show that when the retrieved information is presented to the LLMs in the form of complete TE examples, relevant to the input sentence, their performance gets closer to the fully trained SOTA models. 

In the next section (Section \ref{sec:related_work}), we first provide an outline of those works that experimented with LLMs for TE before. We then proceed, in Section~\ref{sec:pipeline}, to present a pipeline combining LLMs and KBs for the TE task. The proposed pipeline is tested on two standard TE datasets in Section~\ref{sec:experiments}, and the results of an ablation study are reported in Section ~\ref{sec:ablation}. Finally, the closing remarks are presented in Section \ref{sec:conclusion}.

\section{Related Work}\label{sec:related_work}

Classical {\it end-to-end} fully supervised models currently hold the best performance in the TE task. Starting from the older baseline, \citet{zheng-etal-2017-joint}, which also introduced the revised version of the WebNLG dataset for Natural Language Generation (NLG) that is commonly used, several other architectures based on the bidirectional Recurrent Neural Networks (RNNs) \citep{zeng-etal-2018-extracting,fu-etal-2019-graphrel,zeng-etal-2019-learning} have steadily improved the SOTA over the years. More recently, Transformer-based models achieved a big leap forward in performance, with the recent UniRel model being the current SOTA \citep{tang-etal-2022-unirel}.

With the advent of LLMs, \citet{chia-etal-2022-relationprompt} and \citet{kim2023zeroshot}
tested the use of such models for those TE cases where the availability of examples to train on is low. The first work proposed to use a LLM to generate training examples to finetune a {\it Relation Extractor} model to recognize relations for which labels were not available. The latter work, instead, suggested using relation templates of the form {\it ⟨X⟩ relation ⟨Y⟩} and finetune a LLM to fill out {\it ⟨X⟩} and {\it ⟨Y⟩} with the entities appearing in the~sentence.

\citet{Wadhwa2023RevisitingRE},\citet{wei2023zeroshot}, and \citet{zhu2023llms} investigated the general TE task in both Zero- and Few-Shots settings. These studies proposed different approaches based on LLM prompting. The first work tested the Few-Shots performance of GPT-3~\citep{brown2020language} and Text-to-Text Transfer Transformer (T5)~\citep{raffel2023exploring} under the inclusion of manually-crafted and dataset-dependent contextual information in the prompt. 
The second work proposed to perform TE by sequentially prompting ChatGPT in two stages: asking to individuate the possible relation types first and then extracting the entities participating in each relation. The procedure demonstrated better results than a one-stage approach where the model is prompted to extract the triplet directly.
Finally, the third work, evaluated GPT-3~\citep{gpt3} and GPT-4~\citep{gpt4} on some standard benchmarks in the Zero- and One-Shot settings. However, classical fine-tuned models proved to be superior in the majority of the cases.

In our study, we similarly test the Zero- and Few-Shots capabilities of LLMs in two standard TE datasets that have not been covered by these previous works. In contrast to \citet{Wadhwa2023RevisitingRE} that manually crafted static dataset-specific context to be fed to the LLM, we propose here to dynamically gather contextual information useful for extracting the triplets from a KB. This makes our approach more flexible and less data-dependent, as the KB does not require any manual operation and can be easily switched depending on the need. Also, in contrast to other works, we investigate a wide range of Language Models with varying sizes. This allows us to provide an in-depth analysis of the scaling of the performance, both, from the perspective of the model chosen, and the quality of the contextual KB information included in the prompt.

\begin{figure*}
    \centering
    \begin{subfigure}{0.49\linewidth}
    \centering
    \includegraphics[width=\linewidth]{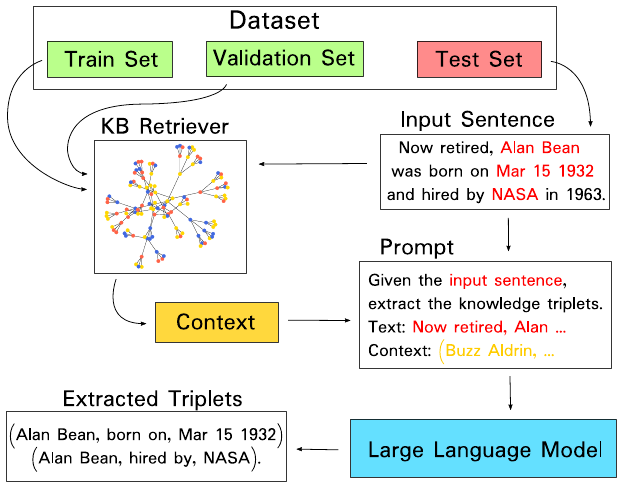}
    \end{subfigure}
    \hfill
    \begin{subfigure}{0.49\linewidth}
    \centering
    \includegraphics[width=\linewidth]{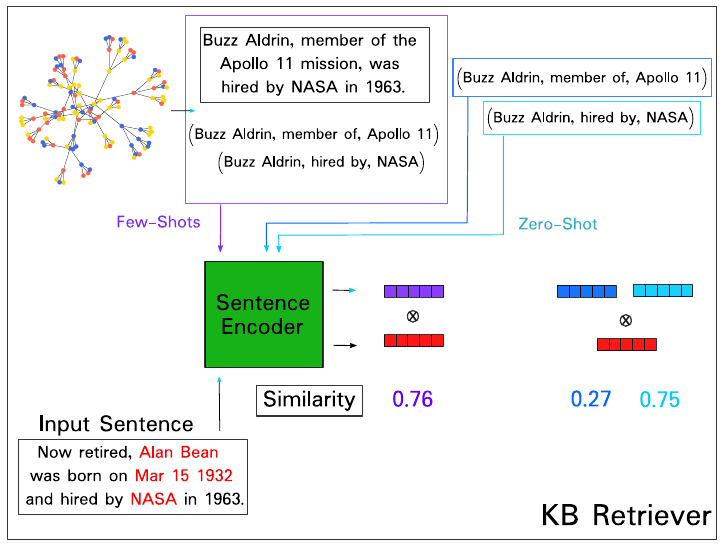}
    \end{subfigure}\\
    \caption{The TE pipeline. Left: illustration of the pipeline. A KB is constructed from the training and validation splits of a given dataset. For each test sentence, the relevant contextual information is retrieved from the KB and included in the prompt for a LLM-based TE. Right: summary of information retrieval from the KB. Either the sentence-triplets pairs or the single triplets alone are encoded by a sentence encoder and compared to the encoding of the input sentence by cosine~similarity.}
    \label{fig:pipeline}
\end{figure*}

\section{The Pipeline}\label{sec:pipeline}

In this section, we provide a detailed illustration of the pipeline used to test the TE capabilities of~LLMs.

\subsection{Task Formulation}

Given a sentence composed of tokens $(t_1,\,t_2,\,\cdots\,,\,t_N)$, the TE task consists of identifying all the relations expressed in it and extracting them in the form of triplets $(s,\,p,\,o)$. Here, $s=(t_i,\,\cdots\,,\,t_{i+n_s})$ and $o=(t_k,\,\cdots\,,\,t_{k+n_o})$ represent a subject and an object of length $n_s$ and $n_o$ tokens, and $p$ is the predicate. Usually, the task is related to a specific KB, {\it i.e.}, a graph of the form $\mathcal{G}=(\mathcal{V},\mathcal{E})$, composed of entities $e\in V$ as vertices and relations $r\in \mathcal{E}$ as directed edges. Therefore, $s$ and $o$ of the sentence correspond to vertices $e_s,e_o\in \mathcal{V}$. The predicate $p$ is mapped to a relation included in the closed set of possible edge types of the KB.

\subsection{LLMs as Triplet Generators}

In order to perform TE, we can prompt LLMs to generate for a given input sentence a sequence of tokens corresponding to the set of entity-relation triplets $\left\{(e_s^i,\,r_p^i,\,e_o^i)\right\}_{i=1}^n$. As demonstrated by \citet{Wadhwa2023RevisitingRE}, \citet{wei2023zeroshot}, and \citet{zhu2023llms}, LLMs are, in principle, able to extract the knowledge triplets contained in a text without a need for task-specific training, under a suitable choice of prompt. In general, successful LLM prompts follow a fixed schema
that provides a detailed explanation of what the task consists of, a clear indication of the sentence to process, and some hints or examples for the desired result.

In this work, we tested the use of three different prompts: a simple baseline and two slight variations of it. 

However, preliminary testing in TE showed no significant difference in the F1 scores among them.
Therefore, we opted for using only the base prompt reported in Figure \ref{fig:base_prompt} in the main experiments. The details of the prompts tested and their results can be found in Appendix \ref{sec:appendix}.

\begin{figure}[b!]
    \centering
    \includegraphics[width=\linewidth]{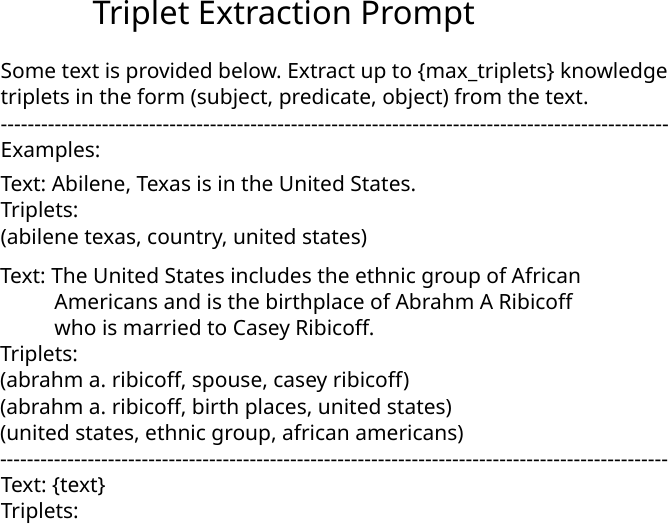}
    \caption{The base prompt we experimented with. At inference time the $\{text\}$ and $\{max\_triplets\}$  variables are substituted with the sentence to process, respectively, the maximum number of triplets found in a sentence in the corresponding dataset.}
    \label{fig:base_prompt}
\end{figure}

\subsection{KB-aided Triplet Extraction}\label{sec:retriever}

In order to support LLMs in the TE task, we propose the pipeline illustrated in Figure \ref{fig:pipeline}. The pipeline augments the LLM with external KB information. In detail, for each input sentence, relevant context information contained in the KB is retrieved and attached to the LLM prompt described above. The context-enriched prompt is then fed to the LLM for the knowledge triplet generation. 

We prepare the information coming from the KB in two different forms: either as simple context~triplets 
\begin{equation}
    T_c=\left\{(e_s^i,\,r_p^i,\,e_o^i)\right\}_{i=1}^{N_{KB}}\in \mathcal{G}\,,
\end{equation}
or as sentence-triplets pairs 
\begin{equation}
    E_c = \left\{(S_c^i,\,T_c^i)\right\}_{i=1}^{N_{KB}}\,.
\end{equation}
The latter provides factual examples of triplets to be extracted for specific sentences. Note that we indicate with $N_{KB}$ the number of triplets, respectively, sentence-triplets pairs retrieved from the~KB. 

In the first case, augmentation is achieved by simply attaching the retrieved triplets $T_c$ as an additional ``Context Triplets'' argument to the base prompt reported in Figure~\ref{fig:base_prompt}. For the second approach, instead, we substitute the two static examples provided in the base prompt, with the input relevant examples $E_c$ retrieved from the~KB.

The relevant context information to build the $T_c$ triplets set for each input sentence is retrieved as follows. Given the KB, we isolate all the triplets $(e_s^i,r_p^i,e_o^i)\in \mathcal{G}$ contained therein, and store them in a node-based vector store index \cite{llamaindex}. In detail, each node of this index corresponds to one and only one of the triplets and stores the embedding obtained by running a small-scale sentence encoder, MiniLM~\citep{wang2020minilm}, on the corresponding $(subject,\,predicate,\,object)$ string. In the first approximation, this should be enough to provide a meaningful embedding for each triplet.\footnote{Other more sophisticated techniques can be imagined to improve the representation and, therefore, the retrieval. However, as the simple embedding method proved to perform well in our case, we did not investigate further.} During inference (\textit{i.e.}, TE), we first encode the input sentence using the MiniLM. This is followed by comparing the obtained sentence embedding with all the triplet embeddings contained in the index to retrieve the top $N_{KB}$ most similar triplets to the input sentence. Out of this $N_{KB}$-dimensional sample, we further select the first two triplets\footnote{In case only one triplet is present in the sample for a specific relation type, we only include that one.} for each relation type present in the sample. This is done to obtain a more diverse set of context triplets with a more homogeneous distribution over the relations. In some cases, indeed, the risk of obtaining a highly biased distribution towards a specific relation type exists, which is sub-optimal for those sentences that contain several different~relationships.

Note that a similar procedure can be followed to prepare the $E_c$ examples set. However, in this case, the focus will be shifted to the example sentences we wish to include. Namely, each node of the vector store index is going to consist both of the example sentence and the KB triplets to be extracted from it. Then, the embedding vector is obtained by running the sentence encoder on either, the example sentence alone, or the sentence and triplets combined. As before, at inference time the top $N_{KB}$ most similar (sentence, triplets) pairs to the input sentence are retrieved and included in the prompt as Few-Shots examples.

\section{Experiments}\label{sec:experiments}

In this section, we first provide details about the datasets and models we tested. This is followed by the presentation of the main results for the TE~task. A summary of the global results is shown in Figure~\ref{fig:violin_plots}, where the distribution of the TE F1 scores over the tested models is plotted as violin plots for the three different settings discussed in the upcoming sections. 

\begin{figure}
    \centering
    \begin{subfigure}{0.49\linewidth}
    \centering
    \includegraphics[width=1.14\linewidth,trim={0cm 0cm 0cm 0cm},clip]{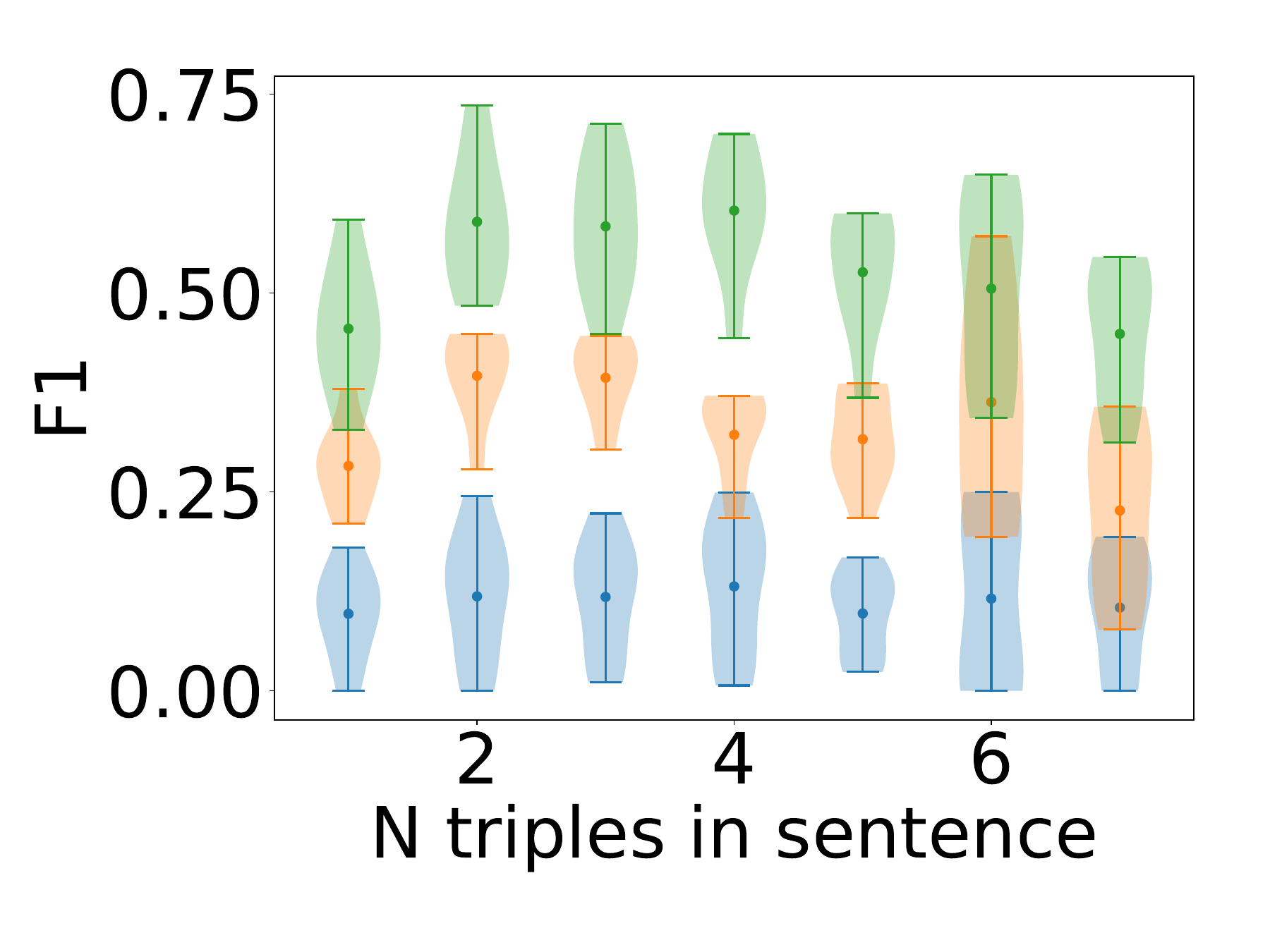}
    \end{subfigure}
    \hfill
    \begin{subfigure}{0.49\linewidth}
    \centering
    \includegraphics[width=0.9\linewidth,trim={6.4cm 0cm 0cm 0cm},clip]{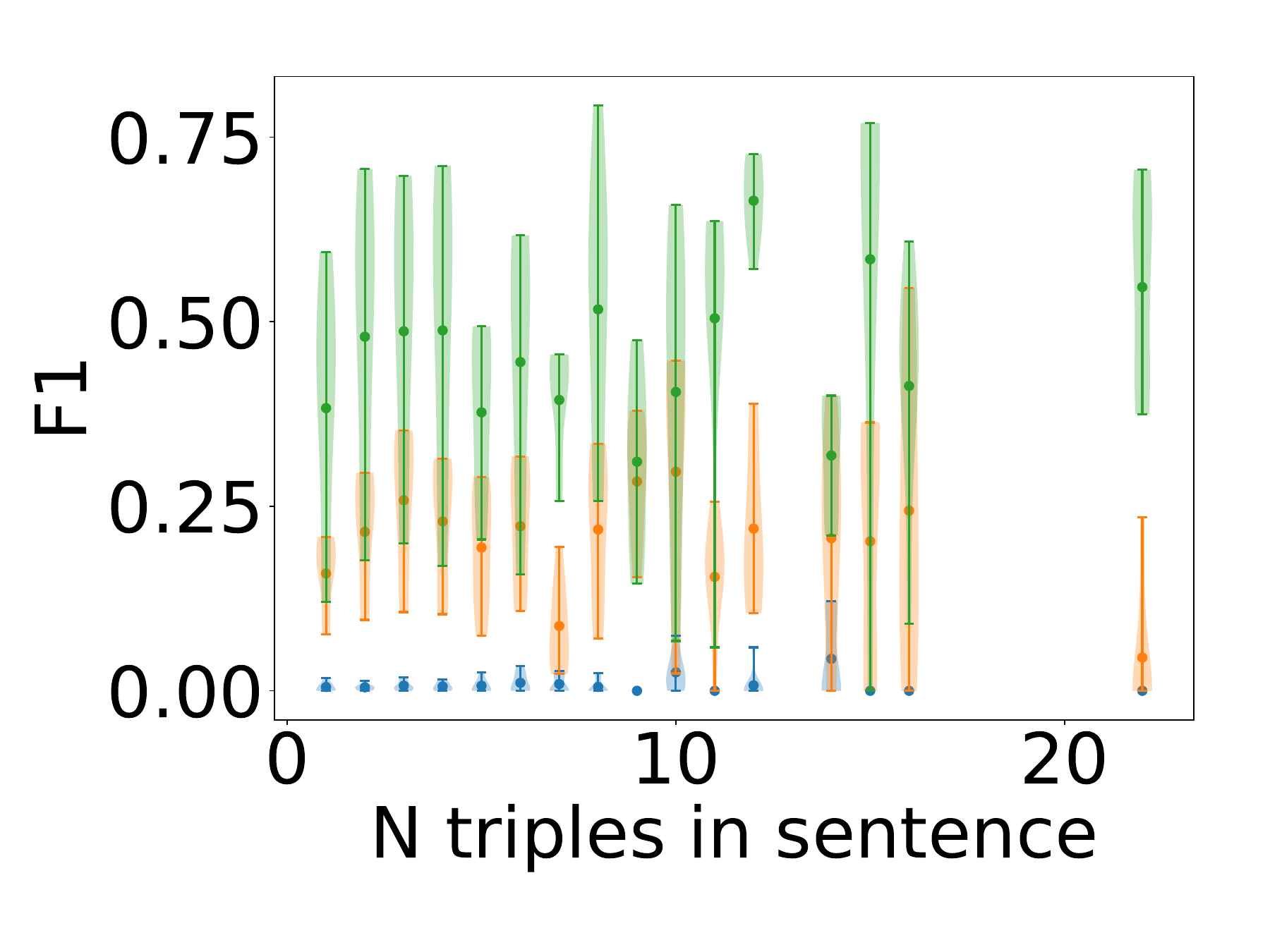}
    \end{subfigure}\\
    \caption{Global summary of the performance obtained by all the models tested in the three settings analyzed in Section~\ref{sec:experiments}: static 2-Shots (blue), 0.5-Shots (orange), 5-Shots (green). The distribution of the micro-averaged F1 across the LLMs is plotted against the number of triplets contained in a sentence for WebNLG (left panel) and NYT (right panel).}
    \label{fig:violin_plots}
\end{figure}

\subsection{Datasets and Models}\label{sec:data}
\setlength{\tabcolsep}{3pt}
\begin{table}[]
\small
    \centering
    \begin{tabular}{l r r r r r r}
          &  Train & Validation & Test & Relations & Max & Avg \\
          \hline
         WebNLG & 5,019 & 500 & 703 & 171 & 7 & 2.29\\
         NYT & 56,195 & 5,000 & 5,000 & 24 & 22 & 1.72
    \end{tabular}
    \caption{Statistics of the WebNLG and NYT datasets. The number of training, validation, and testing sentences is reported, together with the number of relations types in the dataset and the maximum and average number of triplets contained in a sentence.}
    \label{tab:datasets}
\end{table}

\begin{table}[]
\small
    \begin{tabular}{l r r}
          &  Parameters [B] & Context \\
          \hline
         GPT-2~\citep{gpt2} & $0.1\;\vert\;1.5$ & 1,024 \\
         Falcon~\citep{falcon} & $7\;\vert\;40$ & 2,048 \\
         LLaMA~\citep{llama} & $13\;\vert\;65$ & 2,048 \\ 
         GPT-3.5~\citep{gpt3} & 175* & 4,096 \\
         GPT-4~\citep{gpt4} & 1,760* & 8,192 \\
    \end{tabular}
    \caption{The number of parameters (in billions [B]) and context window size of the selected LLMs. We indicate by * the numbers that are not officially confirmed.}
    \label{tab:models}
\end{table}

In order to test the TE capabilities of a selected set of LLMs (see Table \ref{tab:models} for their comparison), we experimented with two standard benchmarks for the TE task: the aforementioned WebNLG~\citep{webnlg} and the New York Times (NYT)~\citep{nyt} dataset (see Table~\ref{tab:datasets} for their basic statistics). The former was initially proposed as a benchmark for the NLG task, but has been successively adapted to the TE task and included in the WebNLG challenge~\citep{webnlg_challenge}. As the revision provided by~\citet{zheng-etal-2017-joint} appears to be the most widely used in the literature, we decided to run our tests on that particular version of WebNLG. The NYT benchmark is a dataset created by distant supervision, aligning more than 1.8 million articles from the NYT newspaper with the Freebase KB. 
For each dataset, we used the training and validation splits to build the corresponding KB following the procedure outlined in Section~\ref{sec:retriever}. 

We selected the LLMs reported in Table~\ref{tab:models} for testing. 
We ran locally the GPT2, Falcon and LLaMA models in their 8-bit quantized version provided by the HuggingFace~\citep{huggingface} library. 
For the OpenAI models, that we ran through the OpenAI API (GPT-3.5 and GPT-4), we were not able to find any information regarding the precision they used. The temperature was set to $\tau=0.1$ for all the experiments. We experimented with higher temperatures but observed that they were detrimental to the TE performance of the model.\footnote{The temperature parameter $\tau$ controls the shape of the Boltzmann distribution from which the generated tokens are sampled. Higher $\tau$ values lead to more flat distributions with heavier tails and therefore, a larger probability of sampling uncommon tokens. Hence, $\tau$ might be regarded as tuning the ``creativity'' of the model: with a large $\tau$ the same prompt could result in widely differing LLM outputs. However, for the TE task, ``creativity'' is not of relevance, as there usually exists only a unique solution.}

Note that for Falcon and LLaMA LLMs, we also explored their {\it instructed} counterparts, {\it i.e.}, models that were fine-tuned for chat applications, either through Reinforcement Learning with Human Feedback (RLHF)~\cite{christiano2023deep} or supervision from other LLMs ~\cite{alpaca}. However, as the instructed models always performed on par, or worse, in our tests, we decided to present the base variants. In the case of GPT-3.5 and GPT-4, we used the instructed versions, as at the time being Open AI only provides the base model for GPT-3 ({\it text-davinci-002}).

We made use of the LlamaIndex~\citep{llamaindex}, LangChain~\citep{langchain} and HuggingFace transformers~\cite{huggingface} python libraries for the implementation of the pipeline.

\renewcommand*{\arraystretch}{1.1}
\setlength{\tabcolsep}{3pt}
\begin{table}[]
\small
    \centering
    \begin{tabular}{|c | c | c|}
    \hline
    Model & WebNLG & NYT \\
      \hline
    NovelTagging~\citep{zheng-etal-2017-joint} & 0.283 &  0.420 \\ 
    CopyRE~\citep{zeng-etal-2018-extracting}, & 0.371 & 0.587 \\
    GraphRel~\citep{fu-etal-2019-graphrel} & 0.429 & 0.619 \\
    OrderCopyRE~\citep{zeng-etal-2019-learning} & 0.616 & 0.721 \\
    UniRel~\citep{tang-etal-2022-unirel} & \textbf{0.947} & \textbf{0.937} \\
    \hline
    \end{tabular}
    \caption{Micro-averaged F1 of some finetuned models selected from the literature.}
    \label{tab:finetuned}
\end{table}

\subsection{Zero- and 2-Shots without the KB}\label{sec:two-shots}

\renewcommand*{\arraystretch}{1.2}
\setlength{\tabcolsep}{1.5pt}
\begin{table}[]
\small
    \centering
 \begin{tabular}{| c  c | c c | c c |}
 \hline
 \multicolumn{2}{|c|}{Model} & \multicolumn{2}{c}{WebNLG} & \multicolumn{2}{|c|}{NYT} \\
 & & 0-Shot & 2-Shots & 0-Shot & 2-Shots \\
 \hline
\multirow{2}{*}{GPT-2} 
 & base & 0.000 & 0.006 & 0.000 & 0.000 \\
 & xl & 0.000 & 0.037 & 0.000 & 0.000\\
 \hline
 \multirow{2}{*}{Falcon} 
 & 7b & 0.000 & 0.066 & 0.000 & 0.002 \\
 & 40b & 0.021 & 0.158 & 0.000 & 0.007 \\
 \hline
 \multirow{2}{*}{LLaMA} 
 & 13b & 0.006 & 0.129 & 0.000 & 0.002 \\
 & 65b  & \textbf{0.041} & \textbf{0.219} & 0.000 & \textbf{0.017}\\
 \hline
 \multirow{2}{*}{OpenAI} 
 & GPT-3.5 & 0.000 & 0.144 & 0.000 & 0.008 \\
 & GPT-4 & 0.007 & 0.156 & 0.000 & 0.007\\
 \hline
 \end{tabular}
    \caption{Zero and 2-Shots micro-averaged F1 performance of the LLMs tested with the prompt of Figure~\ref{fig:base_prompt} and without any context coming from the KB.}
    \label{tab:zero-shot}
\end{table}

As a baseline, we test the Zero- and 2-Shots capabilities of the LLMs without any additional information supplemented from a KB. 
 
As described in Section \ref{sec:pipeline}, we prompt the LLM with the base prompt of Figure~\ref{fig:base_prompt} to extract all the triplets for a sentence in the form (subject, predicate, object). In particular, for the 2-Shots settings, two standard examples are included in the prompt but not changed over the different sentences ({\it c.f.} Figure~\ref{fig:base_prompt}).

In general, the LLMs queried by the base prompt do not seem capable of performing well in the TE task (Table~\ref{tab:zero-shot} summarizes the performance obtained by the LLMs). The two static examples included in the 2-Shots setting help to clarify the task and improve substantially the performance over the Zero-Shot. However, all models struggle to achieve the performance of the classical baseline NLP models, shown in Table~\ref{tab:finetuned}. The sole exception is the LLaMA 65B model that achieves an F1 score close to the one obtained by \citet{zheng-etal-2017-joint} in the WebNLG dataset with 2-Shots. In particular, the NYT benchmark appears to be challenging for LLMs as they have difficulties even reaching a mere 1\% F1 score. This discrepancy in performance between the datasets could potentially be explained as follows: In contrast to the WebNLG dataset, which features more linear and simple sentences, NYT articles often possess a quite complex structure, with several subordinate clauses and implicit relations. In particular, the triplet labels of the NYT dataset often cover only a subset of the actual relations found in the sentence. Therefore, without training examples available, LLMs cannot infer which relations are and are not supposed to be extracted.

\subsection{Zero-shot with KB Triplets (0.5-Shots)}\label{sec:two-shots+triplets}

\renewcommand*{\arraystretch}{1.2}
\setlength{\tabcolsep}{1.5pt}
\begin{table}[]
\small
    \centering
 \begin{tabular}{| c  c | c c | c c |}
 \hline
 \multicolumn{2}{|c|}{Model} & \multicolumn{2}{c}{WebNLG} & \multicolumn{2}{|c|}{NYT} \\
 & & 0.5-Shot & 5-Shots & 0.5-Shot & 5-Shots \\
 \hline
\multirow{2}{*}{GPT-2} 
 & base & 0.249 & 0.430 & 0.175 & 0.375 \\
 & xl & 0.297 & 0.517 & 0.193 & 0.448\\
 \hline
 \multirow{2}{*}{Falcon} 
 & 7b & 0.381 & 0.567 & \textbf{0.250} & 0.519 \\
 & 40b & 0.345 & 0.615 & 0.226 & 0.547 \\
 \hline
 \multirow{2}{*}{LLaMA} 
 & 13b & 0.374 & 0.609 & 0.247 & 0.582 \\
 & 65b  & 0.377 & \textbf{0.677} & 0.243 &\textbf{0.647}\\
 \hline
 \multirow{3}{*}{OpenAI} 
  & text-davinci-002 & \textbf{0.403} & 0.491 & 0.144 & 0.418 \\
 & GPT-3.5 & 0.336 & 0.520 & 0.088 & 0.184 \\
 & GPT-4 & 0.394 & 0.510 & 0.096 & 0.151\\
 \hline
 \end{tabular}
    \caption{0.5 and 5-Shots micro-averaged F1 performance of the LLMs tested with the prompt of Figure~\ref{fig:base_prompt} augmented with $N_{KB}=5$ triplets, respectively, sentence-triplets pairs retrieved from the KB.}
    \label{tab:few-shots}
\end{table}

If we supplement the LLMs with context triplets retrieved from the KB, as described in Section \ref{sec:retriever} and illustrated in Figure \ref{fig:pipeline}, the performance of the LLM in the TE task increases substantially (see Table \ref{tab:few-shots}). 
We refer to this setting where only a set of context triplets, but no example sentence, is provided to the model as 0.5-Shots. The additional triplets hint at which relations and entities the LLM should expect, but they do not give any indication of which sentence pattern they could arise~from.

In this case, the smallest model we tested, namely GPT-2 {\it base}, is competitive with the LLaMA 65B model without context triplets, both, for the WebNLG and the NYT dataset. Furthermore, the bigger models (Falcon, LLaMa, OpenAI) perform better or on par with some of the classical NLP baselines for the WebNLG dataset given in Table~\ref{tab:finetuned}. 

Even so for the NYT dataset a large improvement is obtained under the addition of context triplets, all the LLMs are not able to reach scores competitive with the classical NLP models. The reason behind this might be related to the lower capability of the KB retriever to gather relevant context for NYT ({\it c.f.} Figure~\ref{fig:P_Nkb}) discussed below and to the specific difficulties associated with the NYT dataset discussed in the previous section.

In general, it is interesting to observe that performance with the addition of the context triplets appears to be less dependent on the particular LLM used in case of 0.5-Shot setting. Quite remarkably, the small GPT-2 {\it xl} is able to retain most of the performance of the larger models. This is particularly evident for the NYT dataset, where all the LLMs are not able to perform better than a 25\% F1 threshold. This could be seen as a symptom of the TE accuracy being mainly driven by the added context triplets in this case. Indeed, we also tested this KB triplets augmentation combined with the inclusion of the two static examples used in Section~\ref{sec:two-shots}, but no significant differences were observed. 

A more general remark regards the underwhelming performance of the OpenAI GPT-3.5 and GPT-4 models on the NYT dataset both, in this 0.5-Shots setting, and in the Few-Shots setting discussed in the next section. Our manual inspection of the triplets they provided as an answer suggested that they were less keen to adhere to the entities and relations appearing in the provided KB context, often paraphrasing or reformulating them in a more prolix form that lowered the accuracy. This might be a consequence of the instructed training they had gone through, as discussed in Section~\ref{sec:data}. In contrast, the results provided by the ``text-davinci-002'' model, {\it i.e.}, the base variant of GPT-3, were more in line with all the other~LLMs.

\subsection{Few Shots with KB Sentence-Triplets Pairs}\label{sec:few_shots}

To further aid LLMs in the TE task, we experiment with inclusion in the prompt of input-specific (sentence, triplets) example pairs retrieved from the KB, as detailed in Section \ref{sec:retriever}. Such updated prompts should provide a much stronger signal to the LLM as they not only suggest which entities and relations the LLM should expect, but also which kind of patterns in the sentence correspond to a specific relation. In particular, as it will be discussed in Section~\ref{sec:quality}, the measured train-test overlapping seems to be large for both datasets ({\it c.f.} Figure \ref{fig:P_Nkb}) and, therefore, the updated prompts are likely to include examples of similar sentences. Therefore, performance improvements are expected, and in fact, looking at Table \ref{tab:few-shots}, we see that including 5 of these examples in the prompt makes the LLM competitive with most of the classical baselines reported in Table \ref{tab:finetuned} (except the most recent SOTA from \citet{tang-etal-2022-unirel}).

Interestingly, the performance gap between the two datasets 
narrowed under the updated prompt. In particular, the NYT corpus seems to have become far easier now for the LLMs. As discussed in Section \ref{sec:two-shots}, this dataset consists of sentences with a much more complex structure and more implicit relations. Therefore, having available examples of similarly constructed sentences might have helped the models to more easily identify the correct triplets. 
In addition, in this case, we observed an underwhelming performance for the OpenAI GPT-3.5 and GPT-4 models in the NYT dataset (see Section~\ref{sec:two-shots+triplets}). 

\subsection{Quality of the KB Context}\label{sec:quality}

\begin{figure}
    \centering
\includegraphics[width=\linewidth,trim={5.0cm 11.8cm 5.5cm 11.7cm},clip]{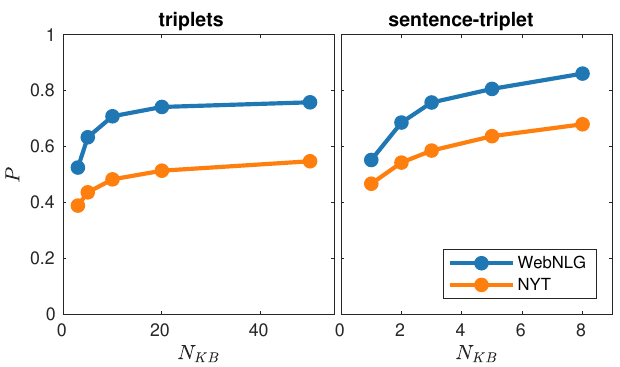}
    \caption{Probability that the correct triplet is present inside the retrieved KB context consisting of (left) triplets alone or (right) sentence-triplet example pairs, plotted against the amount of context gathered, $N_{KB}$.}
    \label{fig:P_Nkb}
\end{figure}

To evaluate the effectiveness of the KB retriever and the quality of the included KB context, we plot in Figure \ref{fig:P_Nkb} the probability of finding the correct triplets with increasing $N_{KB}$, \textit{i.e.}, the solution to the TE task, inside the gathered KB context. Namely, for each test sentence contained in the two datasets, we looped over every labeled triplet and counted the number of times it was contained inside the context provided by the retriever. We repeated this procedure for different values of $N_{KB}$. 

Figure \ref{fig:P_Nkb}(left) suggests that $N_{KB}\sim10-20$ retrieved triplets almost maximize the probability of retrieving a useful context already, as, beyond that, the improvement is only marginal. However, as few as five triplets worked the best in our tests ({\it c.f.} Figure~\ref{fig:f1_vs_Nkb}(left)), with just some exceptions for the sentences that contained a large number of triplets. Probably, a greater number of context triplets retrieved leads to a marginally increased likelihood of including relevant information, but at the cost of a larger dilution. Conversely, as illustrated by Figure~\ref{fig:P_Nkb}(right), for the sentence-triplets augmentation convergence is not reached with $N_{KB}=8$ yet. However, in our experiments, the final TE performance only marginally improved going from 5 to 8 sentence-triplets examples included ({\it c.f.} Figure~\ref{fig:f1_vs_Nkb}(right)). Still, it is interesting to note that LLM performance increases with $N_{KB}$ in this case, providing further evidence that the examples composed of sentence-triplets pairs are much more informative. Adding several of them does not lead to a dilution of useful information, but rather contributes to widening the spectrum of examples the LLM can take inspiration~from.

In general, the probability of providing the correct triplet to the LLM through the context appears to be large: greater than $50\%$ in the majority of the cases, and even approaching the $70\sim80\%$ for the WebNLG dataset. This is symptomatic of substantial overlap that exists between the training, validation, and test splits for both~datasets.

In order to better understand the results obtained by the KB-augmented LLMs, we considered the following simple random TE model: first, we randomly select the number of triplets $n\in\big[1,max\_triplets\big]$ to extract, with $max\_triplets$ indicating the maximum number of triplets contained in a sentence of the dataset. Then, we uniformly sample $n$ triplets out of the retrieved KB~context. 

 Surprisingly, the random model is very competitive with the KB-augmented LLM for small $N_{KB}$ on the WebNLG dataset (see Figure~\ref{fig:f1_vs_Nkb}), and similar results were observed for the NYT dataset. This can be explained by the hand of Figure \ref{fig:P_Nkb}. In detail, 
we infer from the figure that the KB-augmented prompt has a large probability of containing the correct triplets to extract already, therefore, even randomly selecting a subset of them yields a relatively high accuracy. This provides further confirmation that the TE performance is largely driven by the KB~retriever.

\begin{figure}
    \centering
\includegraphics[width=\linewidth,trim={5.0cm 11.8cm 5.5cm 11.7cm},clip]{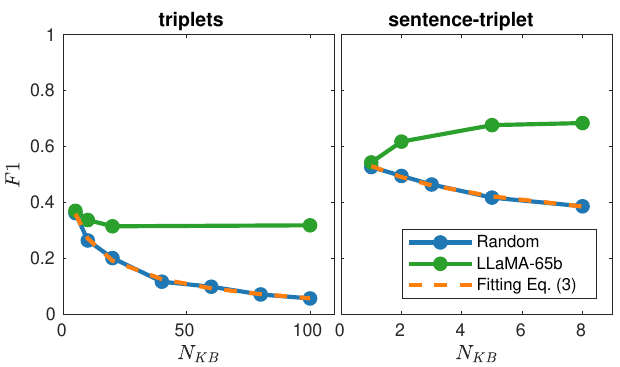}
    \caption{Degradation of the random model performance with the increase of the context information included, $N_{KB}$. The LLaMA-65b, instead, is able to retain most of its performance when more triplets are added (left panel), and sees a significant F1 rise with an increasing number of sentence-triplets pairs (right panel). For reference, we also report the fit of (\ref{eq:random_f1}) as a dashed orange line.}
    \label{fig:f1_vs_Nkb}
\end{figure}

However, the performance of the random model decreases polynomially with $N_{KB}$, as the probability of randomly sampling the correct triplets follows the empirical scaling relation
\begin{equation}\label{eq:random_f1}
    F1_{rand}(N_{KB}) \sim \bigg(\frac{P(N_{KB})}{N_{KB}}\bigg)^n\,,
\end{equation}
with $n$ number of triplets to extract and $P(N_{KB})$ probability of retrieving the correct triplet from the KB (Figure \ref{fig:P_Nkb}).

In contrast, the LLM is able to retain much of its original performance for a larger number of triplets provided ({\it c.f.} Figure \ref{fig:f1_vs_Nkb}(left)) or even improve under the inclusion of more sentence-triplets examples ({\it c.f.} Figure \ref{fig:f1_vs_Nkb}(right)).

\subsection{Ablation Study}\label{sec:ablation}

\begin{figure}
    \centering
\includegraphics[width=\linewidth,trim={5.0cm 11.8cm 5.5cm 11.7cm},clip]{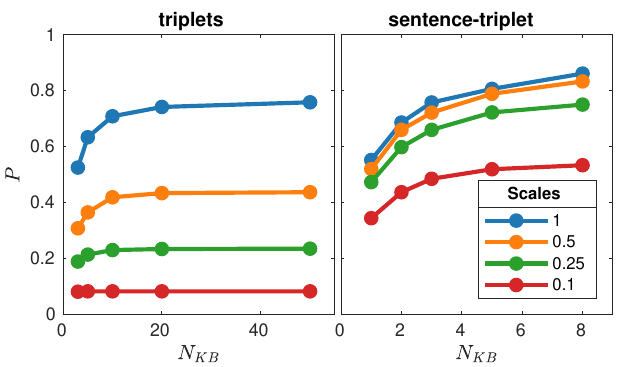}
    \caption{Probability that the correct triplet is present among the retrieved KB context information for the WebNLG dataset, with varying number $N_{KB}$ of context triplets (left panel) or sentence-triplets pairs retrieved (right panel).}
    \label{fig:P_Nkb_scales}
\end{figure}

To further investigate the impact of the additional knowledge retrieved from the KB, we revisit in this section the performance of one of our best performing LLMs, LLaMA-65b. In detail, we construct a scaled-down version of the KB via randomly sampling from the original training and validation splits, keeping only a fraction of the original sentences and triplets. For this reduced KB, the probability of having the correct triplet answer already within the retrieved information is reduced ({\it c.f.} Figure \ref{fig:P_Nkb_scales}). This allows us to evaluate how the accuracy of the model is impacted by the quality of the retrieved data. 

We decided to conduct this test on the WebNLG dataset. As $P(N_{KB})$ for the full-scale KB has been larger than for the NYT dataset, {\it c.f.} Figure~\ref{fig:P_Nkb}, a wider range of values to be explored is allowed. Nonetheless, a preliminary test on the NYT dataset yielded similar results. In Figure \ref{fig:f1_vs_scale} we report the variation of the final F1 score obtained by LLaMA-65b with prompts augmented by $N_{KB}=5$ triplets and sentence-triplets pairs gathered from a KB of different scales $S=0,\,0.1,\,0.25,\,0.5,\,1$. Here, the scale refers to the fraction of left-over data from the original KB. Note that $S=0$ corresponds to the original prompt without any additional information from the KB. In order to have a more general metric, the F1 score is plotted against the probability $P_S(N_{KB}=5)$ of having the correct triplet inside the retrieved data with $N_{KB}=5$ for the different KB sizes. This corresponds to the probability curves of Figure \ref{fig:P_Nkb_scales} evaluated at $N_{KB}=5$. 

We observe that the performance degrades as the probability $P_S(N_{KB})$ shrinks with decreasing $S$, as expected. In particular, the relation appears to be linear:
\begin{align}\label{eq:f1_vs_PNkb_zs}
    F1_{triplets} \sim 0.25 \cdot P_s(N_{KB}=5) + 0.21. \\
    \label{eq:f1_vs_PNkb_fs}
    F1_{\substack{sentence\\-triplets}} \sim 0.55 \cdot P_s(N_{KB}=5) + 0.21.
\end{align}
with measured determination coefficients $r^2=0.98$ and $r^2=0.96$, respectively.
This suggests that there is a strong correlation between the TE capabilities of the model and the quality of the retrieved data. This is a further indication that the TE performance is influenced by the retrieved KB context at least as much as by the chosen LLM. Therefore, a natural question arises: \textit{is it better to use larger models or larger KBs with more reliable information retrievers?}

\begin{figure}
    \centering
    \includegraphics[width=\linewidth,trim={5.0cm 11.8cm 5.5cm 11.7cm},clip]{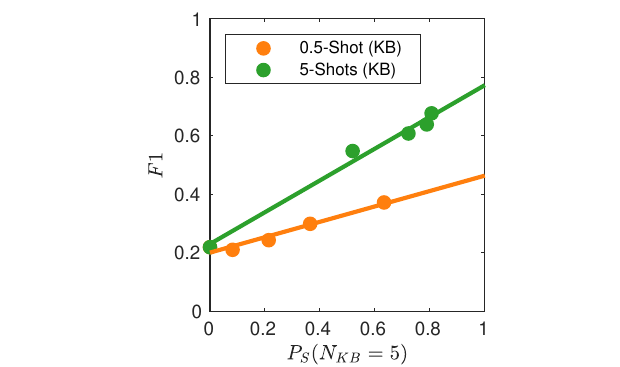}
    \caption{Triplets (orange) and sentence-triplets (green) KB augmented performance of the LLaMA-65b model with different scaled-down versions of the KB built for the WebNLG, $S=0,\,0.1,\,0.25,\,0.5,\,1$. The F1 score is plotted against the probability of retrieving the correct triplet with $N_{KB}=5$ for each $S$ (namely $P(N_{KB}=5)$ for each curve of Figure \ref{fig:P_Nkb_scales}).}
    \label{fig:f1_vs_scale}
\end{figure}

To answer this question, we investigated how the final TE performance scales with the size of the model. 
In Figure \ref{fig:f1_vs_nparams}, the F1 score is plotted against the number of parameters $N_{par}$ in log scale for all the models we tested. The plot includes the results obtained for both the WebNLG and NYT datasets, for all settings considered. We observe that for each of the three settings, the models' performance grows linearly in log scale with respect to their sizes, up to the 2-shot NYT example, {\it c.f.}, related discussions above. However, note that we excluded the GPT-3.5 and GPT-4 0.5- and 5-Shots results obtained on the NYT dataset from the fit, as we consider them to be outliers and not representative, {\it c.f.} Section \ref{sec:two-shots+triplets}.

The scaling in the number of parameters $N_{par}$ in log scale can be approximated by 
\begin{equation}\label{eq:f1_vs_npar}
    F1_{norm} \sim m \cdot \log{N_{par}}.
\end{equation}

The slope parameters of the linear fit for WebNLG are $m=0.0456, 0.0304, \textrm{ and } 0.0871$ for, respectively, 2-Shot, 0.5-Shot(KB), 5-Shots settings, and for the NYT the corresponding parameters are $m=0.0028, 0.0257$ and $0.0906$. The determination coefficients for the WebLNG and NYT datasets are, respectively, $r^2=0.67,\, 0.62$, and $0.97$, and $r^2=0.18,\, 0.7$ and $0.90$. Interestingly, the F1 score increase with the size of the model is steeper for the few-shots prompt ({\it c.f.} Figure~\ref{fig:f1_vs_nparams} right). This suggests that larger models might be more capable in making use of several examples included inside of the prompt.

\begin{figure*}
    \centering
    \includegraphics[width=1\textwidth,trim={0cm 10.5cm 0cm 10.5cm},clip]{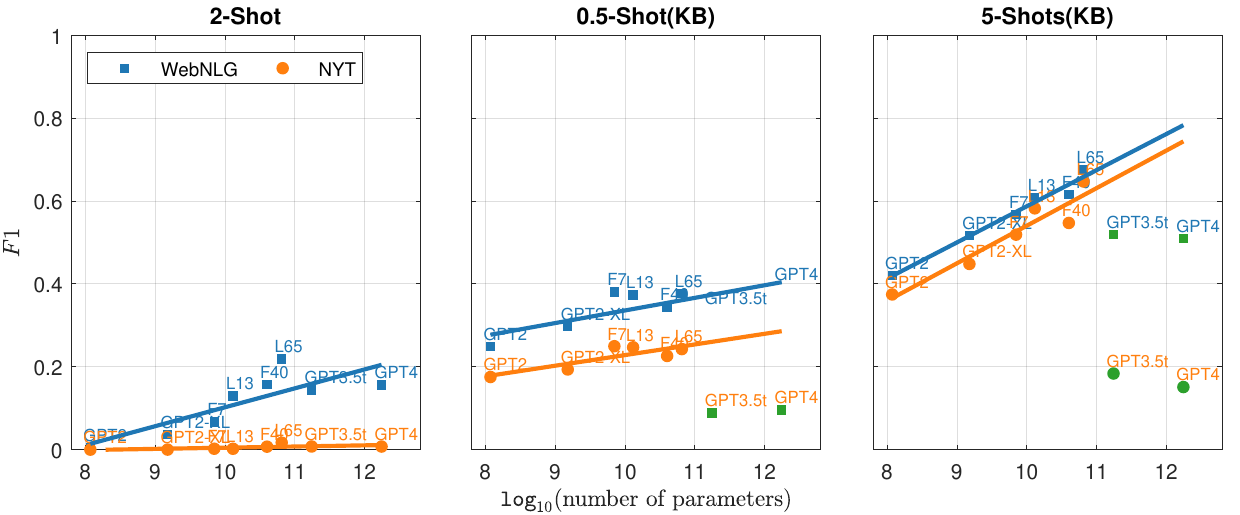}
    \caption{$F1$ score obtained by the tested models, plotted against their corresponding log of number of parameters, for WebNLG (blue) and NYT (orange) in the three settings: 2-shots, 0.5-shots with KB triplets ($N_{KB}=5$), and 5-shots with KB sentence-triplets pairs ($N_{KB}=5$). The outliers (GPT-4 and GPT-3.5 turbo) are shown in green.}
    \label{fig:f1_vs_nparams}
\end{figure*}

Therefore, the $F1$ score and thus the TE accuracy appears to scale linearly with the size of the KB ({\it c.f.} Figure~\ref{fig:f1_vs_scale}), but only logarithmically with the size of the model ({\it c.f.} Figure~\ref{fig:f1_vs_nparams}). This suggests that it could be better to invest resources to improve the quality of the KB and its associated information retriever, rather than in training larger~models.

\section{Conclusion}\label{sec:conclusion}

In this work, a pipeline for Zero- and Few-Shots TE from sentences was presented and tested for various LLMs. We showed that inclusion of KB information into the LLMs prompting can substantially improve the TE performance. In particular, small models were often able to outperform their bigger siblings without access to the additional KB information. Furthermore, with the information from the KB organized as sentence-triplets pair examples relevant to the input sentence, the accuracy of the LLMs improved further. In this setting, the larger LLMs were getting closer to the classical SOTA models and outperformed most of the older~baselines. 

However, even for the largest models, TE remains a challenging task. Currently, LLMs are still no match for SOTA classical {\it end-to-end} models in the two standard benchmark datasets we tested as part of our work, in agreement with~\citet{Wadhwa2023RevisitingRE,wei2023zeroshot,zhu2023llms}. 

The performed investigation of the quality of the retrieved KB context in Section \ref{sec:ablation} raised questions about the generalization capabilities of LLMs for TE. The solution to the TE task was often contained already in the retrieved KB context so that a model that randomly sampled triplets out of the context was competitive with a large LLM, such as LLaMA-65b, in some cases. The observed main advantage of the LLM being its robustness against dilution of the useful information, {\it i.e.}, the LLM was able to identify the relevant triplets among a larger amount of retrieved KB context information. On the contrary, the performance of the random model naturally decayed with increased context.

The competitive performance of the random model, which was also able to match some of the old BiLSTM baselines~\cite{zheng-etal-2017-joint,zeng-etal-2018-extracting,fu-etal-2019-graphrel}, led us to reconsider the generality of the WebNLG and NYT datasets to benchmark TE capabilities. A revised version of the two with reduced training, validation, and test overlapping might be useful to provide a more realistic evaluation of TE models.

We further investigated how the quality of the KB and of the context retrieved from it impacted the performance of the LLaMA-65b model. The experiment gave evidence of the accuracy of the LLM scaling linearly with the probability of finding the solution of the task within the context collected. At the same time, we found that the TE performance improved only approximately logarithmically with the size of the model. This suggests that improving the quality of the KB and the information retriever might be more effective than increasing the modeling power of the LLM for TE.

\section*{Acknowledgements}
Andrea Papaluca was supported by an Australian Government Research Training Program International~Scholarship. Artem Lensky was partially supported by the Commonwealth Department of Defence, Defence Science and Technology Group.

\bibliography{anthology,custom}

\appendix

\section{Appendix}
\label{sec:appendix}

Here we report all the TE prompts that we tested. Figures~\ref{fig:prompt_step} and \ref{fig:prompt_doc} report two variations of the base prompt of Figure~\ref{fig:base_prompt}. The first one implements a Chain-of-Thought~\citep{wei2023chainofthought} approach where multi-step reasoning is enforced. The second tries to provide the LLM with more information about the task, describing in more detail the role of each one of the core components of TE. In Table~\ref{tab:prompt_comparison} the three prompts of Figures~\ref{fig:base_prompt},~\ref{fig:prompt_step}, and~\ref{fig:prompt_doc} are compared for the WebNLG and NYT datasets under the use of two different LLMs, GPT-2~{\it xl}, and LLaMA~65B. The three prompts yield similar micro-averaged F1 scores, with a standard deviation that never exceeds $0.003$. 

Figure~\ref{fig:0.5-prompt} reports the prompt that we used in the 0.5-Shots setting. The prompt consists of a simple adaptation of the base prompt of Figure~\ref{fig:base_prompt} to accommodate for the additional triplets retrieved from the KB.

\begin{figure}[ht]
    \centering
    \includegraphics[width=\linewidth]{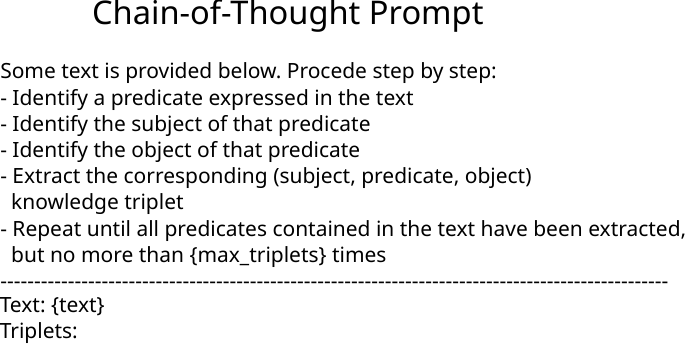}
    \caption{Prompt implementing the Chain-of-Thoughts approach~\cite{wei2023chainofthought}.}
    \label{fig:prompt_step}
\end{figure}

\begin{figure}
    \centering
    \includegraphics[width=\linewidth]{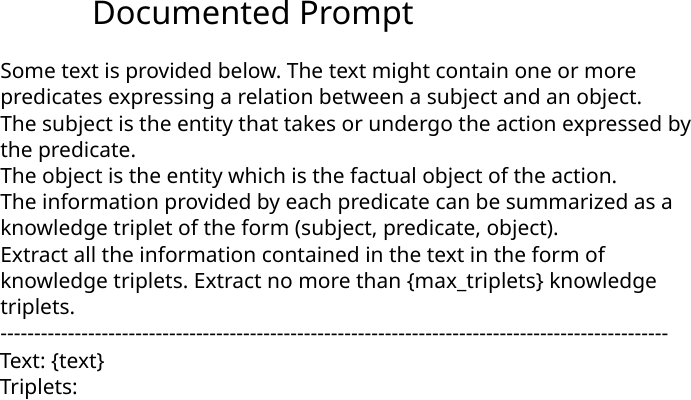}
    \caption{Prompt providing more details about the core components of the TE task, namely, including definitions of subject, object, predicate, and triplet.}
    \label{fig:prompt_doc}
\end{figure}

\begin{figure}
    \centering
    \includegraphics[width=\linewidth]{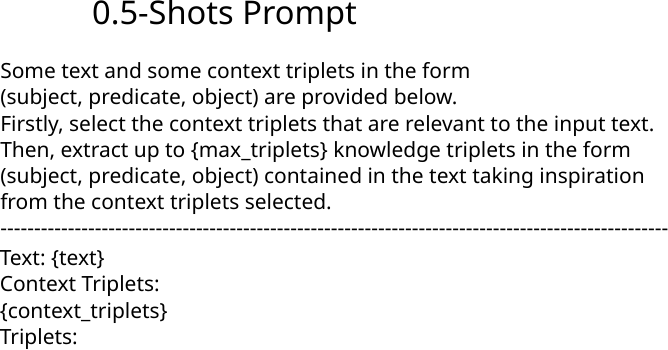}
    \caption{Adaptation of the base prompt found in Figure~\ref{fig:base_prompt} to the 0.5-Shots setting. An additional $\{context\_triplets\}$ argument is included to accommodate for the KB triplets retrieved from the KB.}
    \label{fig:0.5-prompt}
\end{figure}

\renewcommand*{\arraystretch}{1.6}
\begin{table}[]
\small
    \centering
 \begin{tabular}{| c  c | c | c |}
 \hline
 \multicolumn{2}{|c|}{Prompt} & WebNLG & NYT \\
 \hline
\multirow{3}{*}{\rotatebox[origin=c]{90}{GPT-2 xl}} 
 & \textit{base} & 0.037 & 0.0002 \\
 & \textit{documented} &  0.034 & 0.0003\\
 & \textit{chain-of-thought} & 0.039 & 0.0004\\
 \hline
 \hline
  &  &  0.002 & 0.0001 \\
  \hline
  \hline
 \multirow{3}{*}{\rotatebox[origin=c]{90}{LLaMA-65b}} 
 & \textit{base} & 0.219 &  0.017 \\
 & \textit{documented} & 0.213 & 0.012\\
 & \textit{chain-of-thought} & 0.219 & 0.015\\
 \hline
  \hline
  &  & 0.003 &  0.002 \\
  \hline
 \end{tabular}
    \caption{Comparison of the 2-Shots TE micro-averaged F1 performance with the three different prompts of Figures~\ref{fig:base_prompt},~\ref{fig:prompt_step}, and~\ref{fig:prompt_doc}. The standard deviation of the performance across the three prompts is reported below each column. }
    \label{tab:prompt_comparison}
\end{table}

\end{document}